\documentclass[10pt,twocolumn,letterpaper]{article}

\usepackage{cvpr}              

\usepackage{graphicx}
\usepackage{amsmath}
\usepackage{amssymb}
\usepackage{booktabs}
\usepackage{epstopdf}
\usepackage[pagebackref=true,breaklinks=true,letterpaper=true,colorlinks,bookmarks=false]{hyperref}
\usepackage{booktabs}
\usepackage{stfloats}
\usepackage{float}
\usepackage{multirow}
\usepackage{bbding}
\usepackage{bm}
\usepackage{amsfonts,amssymb}
\usepackage{colortbl}
\definecolor{mygray}{gray}{.9}
\definecolor{mypink}{rgb}{.99,.91,.95}
\definecolor{mycyan}{cmyk}{.3,0,0,0}

\usepackage{caption}
\usepackage[capitalize]{cleveref}
\crefname{section}{Sec.}{Secs.}
\Crefname{section}{Section}{Sections}
\Crefname{table}{Table}{Tables}
\crefname{table}{Tab.}{Tabs.}

\def\cvprPaperID{xxxx} 
\def\confName{CVPR}
\def\confYear{2023}

\begin{document}

\title{GALIP: Generative Adversarial CLIPs for Text-to-Image Synthesis}

\author{Ming Tao\textsuperscript{1} \quad 
Bing-Kun Bao\textsuperscript{1,2}\thanks{Corresponding Author} \quad
Hao Tang\textsuperscript{3} \quad 
Changsheng Xu\textsuperscript{2,4,5} \\
\textsuperscript{1}Nanjing University of Posts and Telecommunications \quad  
\textsuperscript{2}Peng Cheng Laboratory \quad \\
\textsuperscript{3}CVL, ETH Zürich \quad 
\textsuperscript{4}University of Chinese Academy of Sciences \\
\textsuperscript{5}NLPR, Institute of Automation, CAS \\
}

\maketitle

\begin{abstract}
Synthesizing high-fidelity complex images from text is challenging.
Based on large pretraining, the autoregressive and diffusion models can synthesize photo-realistic images.
Although these large models have shown notable progress, there remain three flaws.
1) These models require tremendous training data and parameters to achieve good performance.
2) The multi-step generation design slows the image synthesis process heavily.
3) The synthesized visual features are difficult to control and require delicately designed prompts.
To enable high-quality, efficient, fast, and controllable text-to-image synthesis, we propose Generative Adversarial CLIPs, namely GALIP.
GALIP leverages the powerful pretrained CLIP model both in the discriminator and generator.
Specifically, we propose a CLIP-based discriminator.
The complex scene understanding ability of CLIP enables the discriminator to accurately assess the image quality.
Furthermore, we propose a CLIP-empowered generator that induces the visual concepts from CLIP through bridge features and prompts.
The CLIP-integrated generator and discriminator boost training efficiency, and as a result, our model only requires about $3\%$ training data and $6\%$ learnable parameters, 
achieving comparable results to large pretrained autoregressive and diffusion models.
Moreover, our model achieves $\sim$120$\times$faster synthesis speed and inherits the smooth latent space from GAN.
The extensive experimental results demonstrate the excellent performance of our GALIP.
Code is available at \url{https://github.com/tobran/GALIP}.
\end{abstract}

\section{Introduction}
\label{sec:intro}

Over the last few years, we have witnessed the great success of generative models for various applications \cite{wang2020deep,cheng2021fashion}. 
Among them, text-to-image synthesis\cite{hong2018inferring,li2020exploring, cheng2020rifegan, ramesh2021zero, li2019controllable, qiao2019learn, qiao2019mirrorgan, zhu2019dm, yin2019semantics, li2019object,xu2018attngan,liang2020cpgan,liu2020time,wang2022clip,xu2022predict} is one of the most appealing applications. 
It generates high-fidelity images according to given language guidance. 
Owing to the convenience of language for users, text-to-images synthesis has attracted many researchers and become an active research area.

\begin{figure}[t] \small
  \centering
  \includegraphics[width=\linewidth]{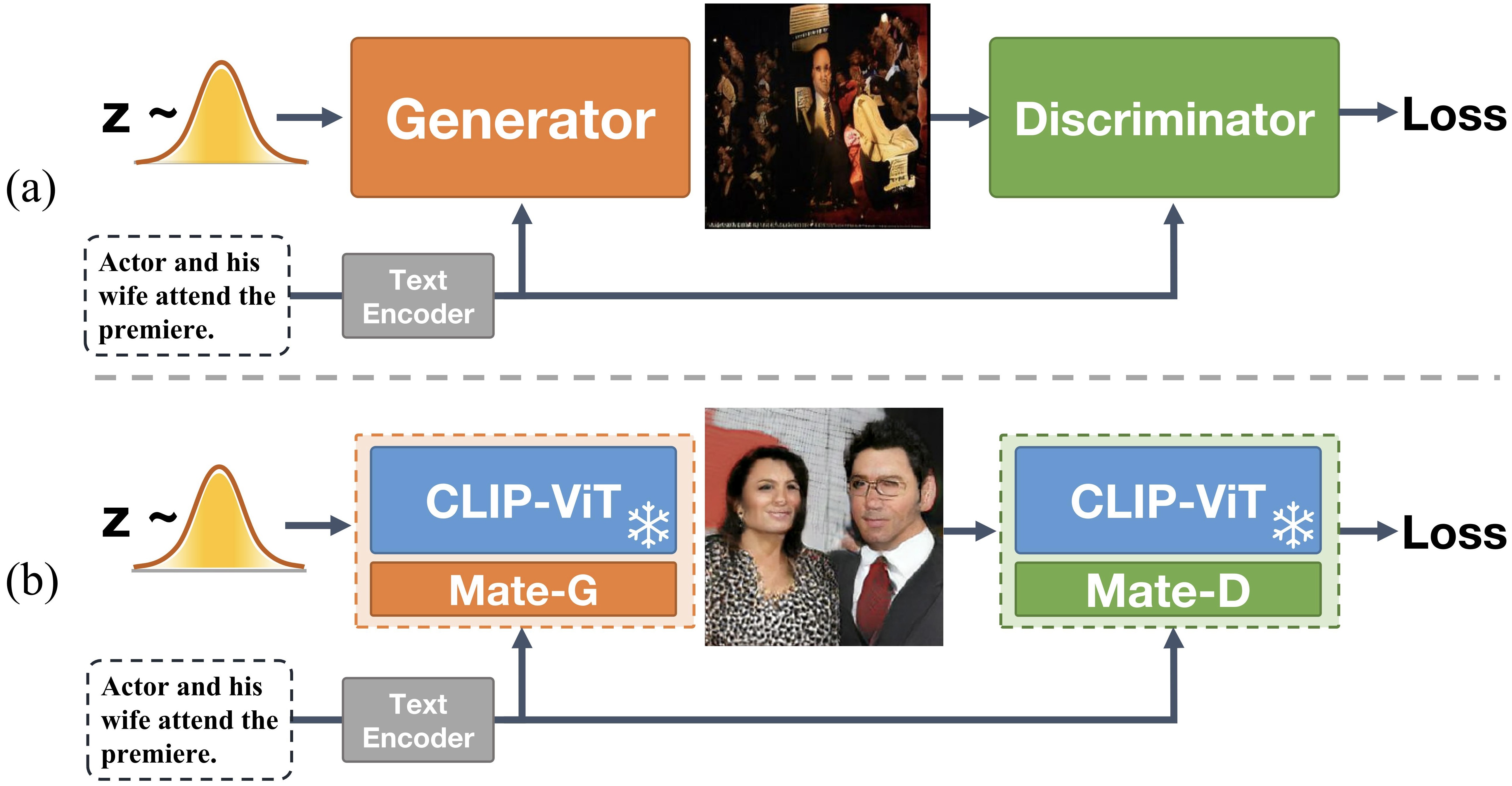}
  \caption{(a) Existing text-to-image GANs conduct adversarial training from scratch. (b) Our proposed GALIP conducts adversarial training based on the integrated CLIP model.}
  \label{fig0}
  \vspace{-0.6cm}
\end{figure}

Based on a large scale of data collections, model size, and pretraining, recently proposed large pretrained autoregressive and diffusion models, \eg, DALL-E~\cite{ramesh2021zero} and LDM~\cite{rombach2022high}, show the impressive generative ability to synthesize complex scenes and outperform the previous text-to-image GANs significantly.
Although these large pretrained generative models have achieved significant advances, they still suffer from three flaws.
First, these models require tremendous training data and parameters for pretraining.
The large data and model size brings an extremely high computing budget and hardware requirements, making it inaccessible to many researchers and users.
Second, the generation of large models is much slower than GANs. 
The token-by-token generation and progressive denoising require hundreds of inference steps and make the generated results lag the language inputs seriously. 
Third, there is no intuitive smooth latent space as GANs, which maps meaningful visual attributes to the latent vector. 
The multi-step generation design breaks the synthesis process and scatters the meaningful latent space.
It makes the synthesis process require delicately designed prompts to control.

To address the above limitations, we rethink Generative Adversarial Networks (GAN). 
GANs are much faster than autoregressive and diffusion models and have smooth latent space, which enables more controllable synthesis.
However, GAN models are known for potentially unstable training and less diversity in the generation \cite{ding2021cogview}. 
It makes current text-to-image GANs suffer from unsatisfied synthesis quality under complex scenes.

\begin{figure*}[t] \small
  \centering
  \includegraphics[width=\linewidth]{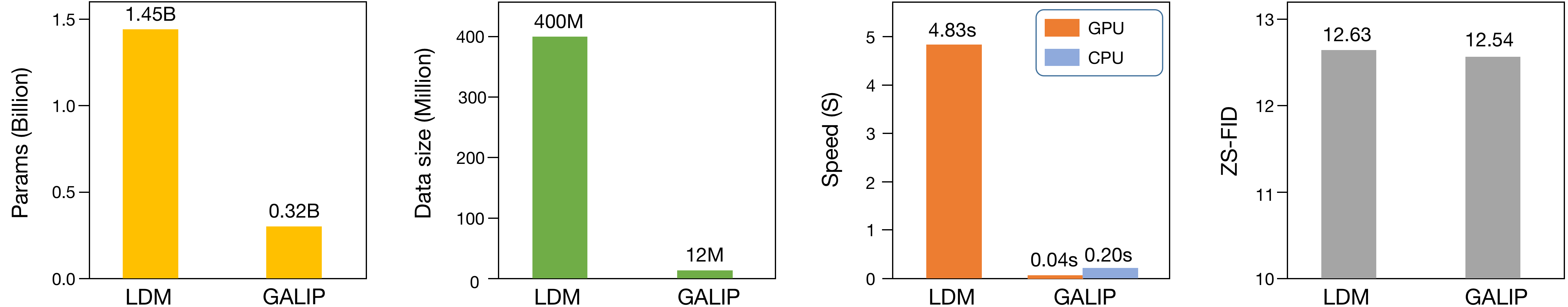}
  \caption{Comparing with Latent Diffusion Models (LDM) \cite{rombach2022high}, our GALIP achieves comparable zero-shot Fr\'echet Inception Distance (ZS-FID) with measly 320M parameters (0.08B trainable parameters + 0.24B frozen CLIP parameters) and 12M training data. Furthermore, our GALIP only requires 0.04s to synthesize one image which is $\sim$120$\times$faster than LDM. Speed is calculated on NVIDIA 3090 GPU and Intel Xeon Silver 4314 CPU.}
  \label{fig1}
  \vspace{-0.4cm}
\end{figure*}

In this work, we introduce the pretrained CLIP \cite{radford2021learning} into text-to-image GANs.
The large pretraining of CLIP brings two advantages.
First, it enhances the complex scene understanding ability.
The pretraining dataset has many complex images under different scenes.
Armed with the Vision Transformer (ViT) \cite{dosovitskiy2020image}, the image encoder can extract informative and meaningful visual features from complex images to align the corresponding text descriptions after adequate pretraining.
Second, the large pretraining dataset also enables excellent domain generalization ability.
It contains various kinds of images, \eg, photos, drawings, cartoons, and sketches, collected from a variety of publicly available sources.
The various images make the CLIP model can map different kinds of images to the shared concepts and enable impressive domain generalization and zero-shot transfer ability.
These two advantages of CLIP, complex scene understanding and domain generalization ability, motivate us to build a more powerful text-to-image model.

We propose a novel text-to-image generation framework named Generative Adversarial CLIPs (GALIP).
As shown in Figure~\ref{fig0}, the GALIP integrates the CLIP model \cite{radford2021learning} in both the discriminator and generator.
To be specific, we propose the CLIP-based discriminator and CLIP-empowered generator.
The CLIP-based discriminator inherits the complex scene understanding ability of CLIP \cite{radford2021learning}.
It is composed of a frozen ViT-based CLIP image encoder (CLIP-ViT) and a learnable mate-discriminator (Mate-D).
The Mate-D is mated to the CLIP-ViT for adversarial training.
To retain the knowledge of complex scene understanding in the CLIP-ViT, we freeze its weights and collect the predicted CLIP image features from different layers.
Then, the Mate-D further extracts informative visual features from collected CLIP features to distinguish the synthesized and real images.
Based on the complex scene understanding ability of CLIP-ViT and the continuous analysis of Mate-D, the CLIP-based discriminator can assess the quality of generated complex images more accurately.

Furthermore, we propose the CLIP-empowered generator, which exerts the domain generalization ability of CLIP \cite{radford2021learning}.
It is hard for the generator to synthesize complex images directly.
Some works employ sketch \cite{gafni2022make} and layout \cite{li2019object,liang2022layout} as bridge domains to alleviate the difficulty.
However, such a design requires additional labeled data.
Different from these works, the excellent domain generalization of CLIP \cite{radford2021learning} motivates us that there may be an implicit bridge domain, which is easier to synthesize but can be mapped to the same visual concepts through the CLIP-ViT.
Thus, we design the CLIP-empowered generator.
It is composed of a frozen CLIP-ViT and a learnable mate-generator (Mate-G).
The Mate-G first predicts the implicit bridge features from text and noise.
Then the bridge feature will be mapped to the visual concepts through CLIP-ViT.
Furthermore, we add some text-conditioned prompts to the CLIP-ViT for task adaptation.
The predicted visual concepts close the gap between text features and target images which enhances the complex image synthesis ability.

As shown in Figure~\ref{fig1}, the proposed GALIP achieves $\sim$120$\times$faster synthesis speed and comparable synthesis ability based on significantly smaller trainable parameters and training data.

Overall, our contributions can be summarized as follows:
\begin{itemize}
    \item We propose an efficient, fast, and more controllable model for text-to-image synthesis that can synthesize high-quality complex images.
    \item We propose the CLIP-based discriminator, which assesses the quality of generated complex images more accurately.
    \item We propose the CLIP-empowered generator, which synthesizes images based on text features and predicted CLIP visual features.
    \item Extensive experiments demonstrate that the proposed GALIP can achieve comparable performance with large pertaining models based on significantly smaller computational costs.
\end{itemize}

\begin{figure*}[t]
  \centering
  \includegraphics[width=\linewidth]{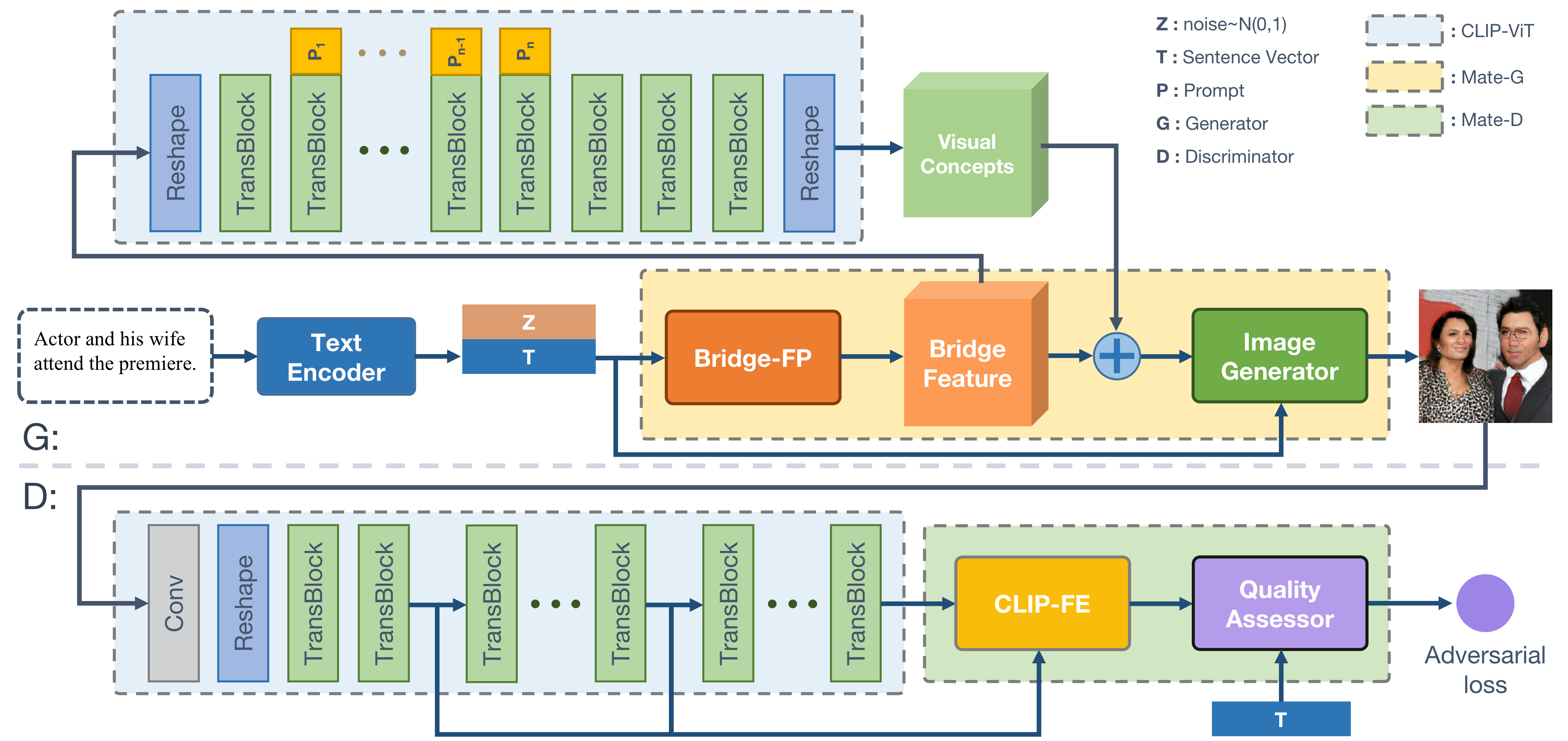}
  \caption{The architecture of the proposed GALIP for text-to-image synthesis. Armed with the CLIP-based discriminator and CLIP-empowered generator, our model can synthesize more realistic complex images.}
  \label{fig2}
  \vspace{-0.4cm}
\end{figure*}

\section{Related Work}

\noindent{\bf Text-to-Image GANs.} 
GAN-INT-CLS \cite{reed2016generative} first adopted conditional GANs to synthesize images from text descriptions.
To enable higher resolution synthesis, the StackGAN\cite{zhang2017stackgan,zhang2018stackgan}, AttnGAN\cite{xu2018attngan}, and DM-GAN \cite{zhu2019dm} stacks multiple generators and discriminators. 
Tao \emph{et al.} \cite{tao2020df} proposed a simpler yet effective text-to-image framework called DF-GAN that enables one-stage high-resolution generation.
LAFITE \cite{zhou2022towards} introduces CLIP text-image contrastive loss for text-to-image training and shows large improvements on CC3M\cite{sharma2018conceptual}. 

\noindent{\bf Text-to-Image Large Models.} 
Recently, large pretrained autoregressive and diffusion models have shown impressive performance on text-to-image synthesis.
DALL-E\cite{ramesh2021zero}, CogView\cite{ding2021cogview}, and M6 \cite{lin2021m6} leverage VQ-VAE \cite{van2017neural} or VQ-GAN \cite{esser2021taming} to tokenize the images into discrete image tokens.
Then they take the word tokens and image tokens together to pre-train a large unidirectional transformer for an autoregressive generation.
Parti \cite{yu2022scaling} proposes a sequence-to-sequence autoregressive model to treat text-to-image synthesis as a translation task.
Cogview2 \cite{ding2022cogview2} employs hierarchical transformers and local parallel autoregressive generation for faster autoregressive image generation.
Some works try to employ the diffusion model \cite{sohl2015deep,dhariwal2021diffusion,ho2020denoising,ho2022cascaded,nichol2021improved} to overcome the slow generation defect of the autoregressive model.
VQ-Diffusion \cite{gu2022vector} combines the VQ-VAE \cite{van2017neural} and diffusion model \cite{ho2022cascaded,nichol2021improved} to eliminate the unidirectional bias and avoids accumulated prediction errors.
GLIDE \cite{nichol2021glide} applies guided diffusion to the problem of text-conditional image synthesis.
DALL-E2 \cite{ramesh2022hierarchical} combines the CLIP representation and diffusion model to make a CLIP decoder.
Latent Diffusion Models (LDM) \cite{rombach2022high} apply the diffusion model in the latent space to enable the training on limited computational resources while retaining image quality.
The particular text-to-image LDM is Stable Diffusion \cite{stablediff}, which is a favorite open-source project and provides an easy-to-use interface.
Imagen \cite{saharia2022photorealistic} introduces the large language model \cite{raffel2020exploring} to provide high-quality text features and proposes an Efficient U-Net for diffusion models.


\section{Generative Adversarial CLIPs }

In this paper, we propose a novel framework for text-to-image synthesis named Generative Adversarial CLIPs (GALIP). 
To synthesize high-quality complex images, we propose: 
(i) a novel CLIP-based discriminator that inherits the complex scene understanding ability of CLIP \cite{radford2021learning} for more accurate image quality assessment.
(ii) a novel CLIP-empowered generator that exerts the domain generalization ability of CLIP \cite{radford2021learning} and induces the CLIP visual concepts to close the gap between text and image features.
In the following of this section, we first present the overall structure of our GALIP. 
Then, we introduce the CLIP-based discriminator and CLIP-empowered generator in detail.

\subsection{Model Overview}

As shown in Figure \ref{fig2}, the proposed GALIP is composed of a CLIP text encoder, a CLIP-based discriminator, and a CLIP-empowered generator. 
The pretrained CLIP text encoder takes the text description and yields a global sentence vector $\bm{T}$.
After the text-encoder is the CLIP-empowered generator and CLIP-based discriminator under the GAN framework.
The CLIP-empowered generator is composed of a frozen CLIP-ViT and a mate generator (Mate-G). 
There are three main modules in the Mate-G, the bridge feature predictor (Bridge-FP), the prompt predictor, and the image generator.
The CLIP-empowered generator has two inputs, the sentence vector $\bm{T}$ encoded from the text encoder and the noise vector $\bm{Z}$ sampled from the Gaussian distribution.
The noise vector ensures the diversity of the synthesized images.
In the CLIP-empowered generator, the sentence vector and noise are first fed into the bridge feature predictor.  
The bridge feature predictor translates the sentence vector and noise to the bridge feature for the CLIP-ViT.
Furthermore, we add several text-conditioned prompts to the transformer blocks (TransBlock) in CLIP-ViT for task adaptation.
Finally, the image generator takes the predicted visual concepts, bridge features, sentences, and noise vectors to synthesize high-quality images.

\begin{figure*}[t]
  \centering
  \includegraphics[width=\linewidth]{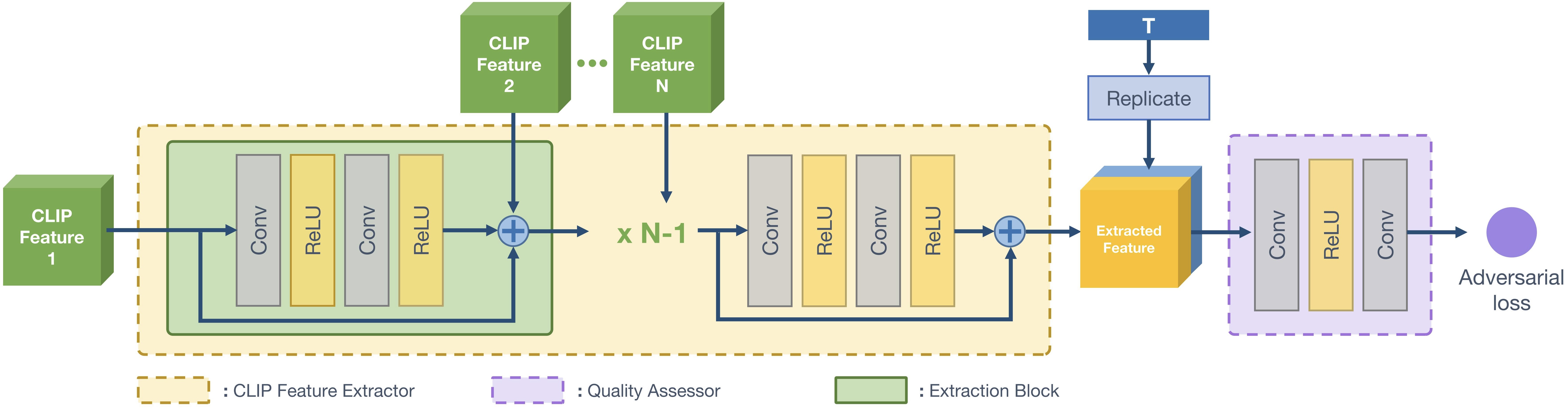}
  \caption{The architecture of the proposed Mate-D for text-to-image synthesis. It further extracts informative visual features from collected CLIP features and assesses the image quality more accurately.}
  \label{fig3}
  \vspace{-0.4cm}
\end{figure*}

The CLIP-based discriminator is composed of a frozen CLIP-ViT and a mate discriminator (Mate-D).
The CLIP-ViT converts images into image feature through a convolution layer and a series of transformer blocks.
The CLIP feature extractor (CLIP-FE) in Mate-D collects the image features from different layers in CLIP-ViT.
Then it further extracts informative visual features from collected CLIP features for the quality assessor.
Lastly, an adversarial loss will be predicted by the quality assessor based on the extracted informative features and sentence vectors.
By distinguishing synthesized images from real ones, the discriminator promotes the generator to synthesize higher-quality images.

\subsection{CLIP-based Discriminator}

In this section, we detailed the proposed CLIP-based discriminator, which is composed of a frozen CLIP-ViT and a Mate-D. 
The CLIP-based discriminator inherits the complex scene understanding ability from the frozen CLIP-ViT.
Furthermore, we propose the Mate-D, which is mated to the CLIP-ViT to further extract informative visual features and distinguish real and synthesized images.
The CLIP-ViT and Mate-D enable the discriminator to assess the quality of generated complex images more accurately. 

As shown in Figure \ref{fig3}, the Mate-D consists of a CLIP-FE and a quality assessor.
To fully utilize the knowledge of complex scene understanding in CLIP-ViT,
the CLIP-FE takes the CLIP image features from multilayers.
There are $N$ CLIP features collected for the CLIP-FE.
We name them CLIP Feature $1$ to $N$, which are collected from shallow to deep layers in CLIP-ViT.
To further extract informative visual features from these CLIP features, we design a CLIP-FE.
It contains a sequence of extraction blocks, and each block contains two convolution layers and two ReLU active functions.
And the extracted image feature is summed with the shortcut and the next CLIP feature.
There are $N-1$ extraction blocks stacked in CLIP-FE.
Since the CLIP feature $N$ is only added to the processed image features in the last extraction block.
To fuse the CLIP feature $N$, we append two convolution layers without the CLIP feature addition behind. 
The CLIP-FE extracts informative visual features for the quality assessor. 
Then the sentence vector is replicated and concatenated with the extracted image features.
An adversarial loss is predicted by two convolution layers to evaluate the image quality. 
Furthermore, to stabilize the adversarial learning process of Mate-D, we apply the matching-aware gradient penalty (MAGP) \cite{tao2020df} on the collected CLIP features and corresponding text features.

Based on the complex scene understanding ability of CLIP-ViT, the CLIP-based discriminator can extract more informative visual features from complex images.
The higher-quality extracted visual features make it easier for the discriminator to detect unreal image parts, which improves the discriminative efficiency, thus prompting the generator to generate more realistic images.

\subsection{CLIP-empowered Generator}
In this section, we detail the proposed CLIP-empowered generator, which is composed of a frozen CLIP-ViT and a Mate-G.
The CLIP-empowered generator exerts the domain generalization ability of the CLIP-ViT.
Furthermore, we propose the Mate-G, which is mated to the CLIP-ViT to induce useful visual features from the CLIP-ViT and generate images from text and induced visual features.
The Mate-G consists of a Bridge Feature Predictor (Bridge-FP), a prompt predictor, a frozen CLIP-ViT, and an image generator (see Figure \ref{fig2}). 
We detail them next.

\begin{figure*}[t]
  \centering
  \includegraphics[width=\linewidth]{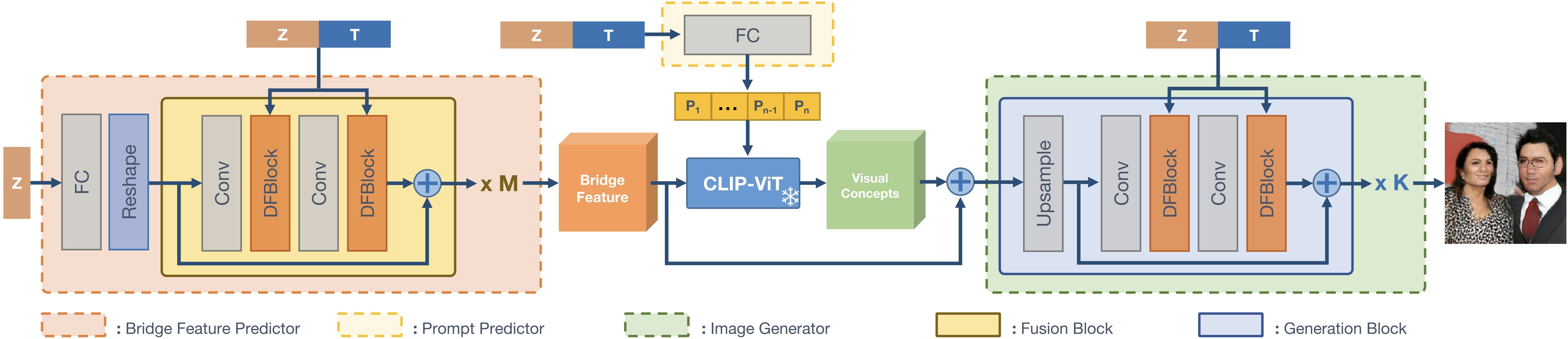}
  \caption{The architecture of the proposed CLIP-empowered generator for text-to-image synthesis. Armed with bridge feature predictor and prompt predictor, it can induce meaningful visual concepts from the frozen CLIP-ViT for image synthesis.}
  \label{fig4}
  \vspace{-0.4cm}
\end{figure*}

\noindent \textbf{Bridge Feature Predictor.} The structure of the Bridge-FP is shown in Figure \ref{fig4}, as highlighted by the red dashed box.
The Bridge-FP consists of an FC (Fully-Connected) layer and $M$ fusion blocks (F-BLKs).
The input noise is fed into the FC layer and reshaped to $(7,7,64)$ as an initial bridge feature.
The initial bridge feature output by the FC layer still contains a lot of noise. 
Therefore, we apply a sequence of F-BLKs to fuse text information and make it more meaningful.
The F-BLK is composed of two convolution layers (Conv) and two deep text-image fusion blocks (DFBlock) \cite{tao2020df}.
The DFBlock has shown its effectiveness in fusing text and image features through stacked affine transformations.
Thus, we adopt it to fuse text features and intermediate bridge features.
There is a shortcut addition in F-BLK for effective information propagation and gradient back-propagation.
Through the Bridge-FP, the sentence and noise vectors will be translated to the bridge feature, which is adjusted to induce meaningful visual concepts from CLIP-ViT.

\noindent \textbf{Prompt Predictor.} The CLIP-ViT is pretrained to predict visual features from image data.
There is a large gap between text and image data.
To alleviate the difficulty of bridge feature translation from text features, we employ prompt tuning \cite{jia2022visual}, which has shown effectiveness on domain transferring for ViT.
We design a prompt predictor, which predicts prompts based on sentence and noise vectors through an FC layer.
The predicted text-conditioned prompts are appended behind the visual patch embeddings in CLIP-ViT.
Furthermore, we find that it is better not to add prompts to the last few layers in CLIP-ViT. 
The last few layers summarize the visual features and output the last image representations. 
The prompt predicted from text and noise in the last few layers may defect its performance.

\noindent \textbf{Image Generator.} The image generator consists of $K$ generation blocks (G-BLKs).
We sum the predicted visual concepts and bridge features through shortcut addition for effective information propagation and gradient back-propagation.
The image generator receives the summed visual features as input and fuses sentence and noise vectors through the DFBlocks \cite{tao2020df} in each G-BLK.
The intermediate image features grow larger during the generation process by the upsample layers.
Finally, the image features are converted into high-resolution RGB images.

\subsection{Objective Functions}

To stabilize the training process of adversarial learning, we employ the hinge loss \cite{zhang2019self} and one-way discriminator \cite{tao2020df}.
Finally, the whole formulation of our GALIP is shown as follows:

\begin{small} 
\begin{equation}
 \begin{aligned}
 L_D = &-\mathbb{E}_{x \sim \mathbb{P}_{r}}[min(0,-1+D(C(x),e))]\\
       &-(1/2)\mathbb{E}_{G(z,e)\sim \mathbb{P}_{g}}[min(0,-1-D(C(G(z,e)),e))]\\
       &-(1/2)\mathbb{E}_{x \sim \mathbb{P}_{mis}}[min(0,-1-D(C(x),e))]\\
       &+k\mathbb{E}_{x \sim \mathbb{P}_{r}}[(\|\nabla_{C(x)}D(C(x),e)\|+\|\nabla_{e}D(C(x),e)\|)^{p}],\\
 L_G = &-\mathbb{E}_{G(z,e)\sim \mathbb{P}_{g}}[{D(C(G(z,e)),e)}]\\
       &-\lambda\mathbb{E}_{G(z,e)\sim \mathbb{P}_{g}}[S(G(z,e),e)], \\
 \end{aligned} 
\end{equation}
\end{small}
where $z$ is the noise vector sampled from Gaussian distribution; $e$ is the sentence vector; $G$ is the CLIP-empowered generator; $D$ is the Mate-D; $C$ is the frozen CLIP-ViT in CLIP-based discriminator; $S$ represents the cosine similarity between the encoded visual and text features of CLIP; $k$ and $p$ are two hyper-parameters of gradient penalty; $\lambda$ is the coefficients of the text-image similarity;$\mathbb{P}_{g}$, $\mathbb{P}_{r}$, $\mathbb{P}_{mis}$ denote the synthetic data distribution, real data distribution, and mismatching data distribution, respectively.

\begin{figure*}[t] \small
  \centering
  \includegraphics[width=\linewidth]{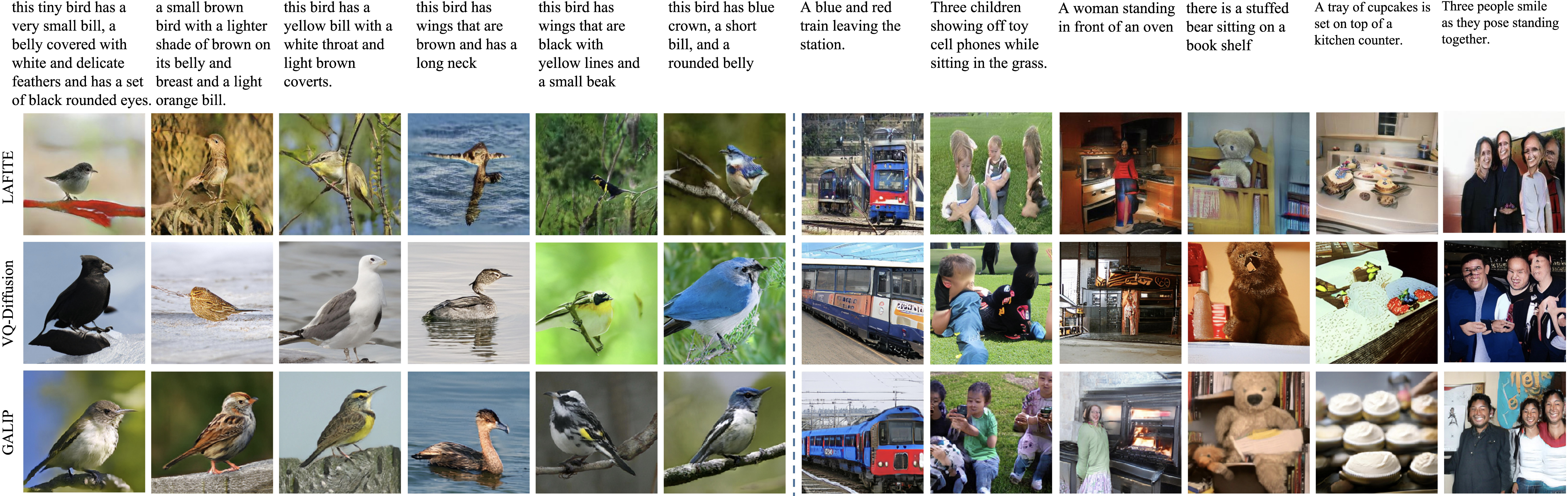}
  \caption{Examples of images synthesized by LAFITE \cite{zhou2022towards}, VQ-Diffusion \cite{gu2022vector}, and our proposed GALIP conditioned on text descriptions from the test set of CUB and COCO datasets.}
  \label{fig5}
  \vspace{-0.4cm}
\end{figure*}

\section{Experiments}

In this section, we introduce the datasets, training details, and evaluation metrics employed in our experiments, then evaluate our proposed GALIP and its variants quantitatively and qualitatively. 

\noindent{\bf Datasets.} We conduct experiments on four challenging datasets: CUB bird \cite{wah2011caltech}, COCO \cite{lin2014microsoft}, CC3M \cite{sharma2018conceptual}, and CC12M \cite{changpinyo2021cc12m}.
For the CUB bird dataset, 
there are 11,788 images belonging to 200 bird species, with each image corresponding to ten language descriptions.
The train and validation splits of the CUB bird dataset are implied as previous works did \cite{zhang2017stackgan, zhang2018stackgan, xu2018attngan, zhu2019dm, tao2020df}.
Since there are various shapes, colors, and postures of birds in the CUB dataset, it is always employed to evaluate the performance of fine-grained content synthesis.
For COCO dataset, 
it contains 80k images for training and 40k images for testing. 
Each image corresponds to 5 language descriptions.
The image in the COCO dataset is complex and always contains multiple objects under different scenes.
The COCO dataset is always employed in recent works to evaluate the performance of complex image synthesis.
For CC3M and CC12M datasets, 
they are two large datasets that contain about 3 and 12 million text-image pairs.
It is always adopted for pretraining and to evaluate the zero-shot performance of the text-to-image model.

\noindent{\bf Training and Evaluation Details.} 
We choose the ViT-B/32 \cite{radford2021learning} model as the CLIP model in our GALIP.
In the CLIP-based discriminator, the CLIP-FE collects the CLIP feature from 2$^{nd}$, 5$^{th}$, 9$^{th}$ layers in CLIP-ViT.
There are two extraction blocks stacked in CLIP-FE.
In the CLIP-empowered generator, the Bridge-FP contains 4 Fusion Blocks, and the image generator contains 6 generation blocks for $224{\times}224$ image synthesis.
The prompt predictor predicts 8 prompts for TransBlocks 2 to 10 in CLIP-ViT.
We conduct some ablation studies on these designs.  
The hyper-parameters of the discriminator $k$ and $p$ are set to 2 and 6 as \cite{tao2020df}.
The hyper-parameters of the generator $\lambda$ are set to 4 for all the datasets.
Furthermore, we employ the Adam optimizer \cite{kingma2014adam} with $\beta_{1}{=}0.0$ and $\beta_{2}{=}0.9$ to train our model. 
According to the two timescale update rule (TTUR) \cite{heusel2017gans}, the learning rate is set to 0.0001 for the generator and 0.0004 for the discriminator. 
Following the previous text-to-image works \cite{xu2018attngan, zhu2019dm, tao2020df, wu2021n}, we adopt the Fr\'echet Inception Distance (FID) \cite{heusel2017gans} and CLIPSIM \cite{wu2021n} to evaluate the image fidelity and text-image semantic consistency.
All GALIP models are trained on 8$\times$3090 GPUs.
We train our GALIP for 0.5, 1.5, 2, and 3 days on CUB, COCO, CC3M, and CC12M datasets, respectively.

\begin{table}[t] \small
\centering
\caption{The results of FID and CLIPSIM (CS) compared with the state-of-the-art methods on the test set of CUB and COCO.}
\begin{tabular}{l|c|c|c|c}
\toprule
\multirow{2}*{Model}                 & \multicolumn{2}{c|}{CUB}                            & \multicolumn{2}{c}{COCO}             \\ 
\cline{2-5}   
                                     & FID $\downarrow$        & CS $\uparrow$             & FID $\downarrow$          & CS $\uparrow$   \\ \midrule
DM-GAN  \cite{zhu2019dm}             & 16.09                   & -                         & 32.64                     & -                  \\
XMC-GAN   \cite{zhang2021cross}      & -                       & -                         & 9.30                      & -                  \\
DAE-GAN   \cite{ruan2021dae}         & 15.19                   & -                         & 28.12                     & -               \\
DF-GAN    \cite{tao2020df}           & 14.81                   & 0.2920                    & 19.32                     & 0.2972           \\ 
LAFITE   \cite{zhou2022towards}      & 14.58                   & 0.3125                    & 8.21                      & 0.3335         \\ 
VQ-Diffusion \cite{gu2022vector}     & 10.32                   & -                         & 13.86                     & -             \\ \bottomrule
GALIP (Ours)                         & \textbf{10.08}          & \textbf{0.3164}           & \textbf{5.85}             & \textbf{0.3338}   \\ \bottomrule
\end{tabular}
\label{table1}
\vspace{-0.4cm}
\end{table}

\subsection{Quantitative Evaluation}
To evaluate the performance of our GALIP, we compare the proposed model with several state-of-the-art methods \cite{zhu2019dm,zhang2021cross,ruan2021dae,zhou2022towards,tao2020df,gu2022vector}, which have achieved impressive results in text-to-image synthesis. The results are shown in Table \ref{table1}.
Compared with other leading models, our GALIP has a significant improvement on both CUB and COCO datasets.
Especially, compared with the recently proposed LAFITE \cite{zhou2022towards}, which employs CLIP text-image contrastive loss for text-to-image training, 
our GALIP decreases the FID metric from 14.58 to 10.08 and improves the CLIPSIM (CS) from 0.3125 to 0.3164 on the CUB dataset. 
Furthermore, our GALIP decreases the FID of COCO from 8.21 to 5.85 significantly.
Compared with VQ-diffusion \cite{gu2022vector}, which adopts diffusion models for text-to-image synthesis, 
our GALIP also decreases FID from 10.32 to 10.08 on the CUB dataset and decreases the FID of COCO from 13.86 to 5.85 remarkably.
The quantitative comparisons on CUB and COCO datasets demonstrate that our GALIP is more effective in synthesizing high-fidelity images, especially for complex image generation.

\begin{figure*}[t] \small
  \centering
  \includegraphics[width=\linewidth]{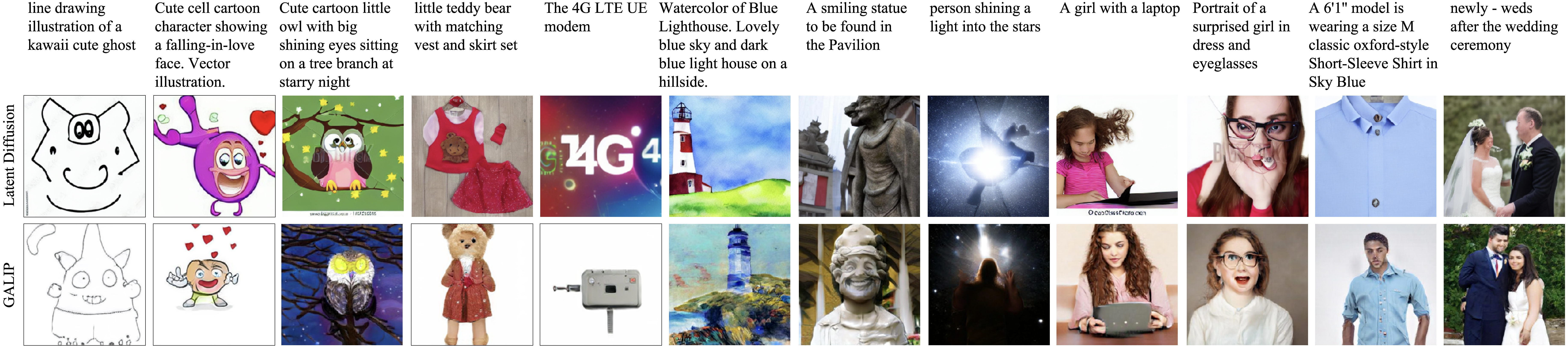}
  \caption{Text-to-Image samples from GALIP (CC12M) and Latent Diffusion (LAION-400M) \cite{rombach2022high,stablediff}. We sample 16 images from each given text description, and randomly select one as the final generation result}
  \label{fig6}
  \vspace{-0.4cm}
\end{figure*}

\begin{table}[t] \small
\centering
\caption{We compare the performance of large pretrained autoregressive models (AR), diffusion models (DF), and GANs under zero-shot setting on the COCO test dataset.}
\resizebox{1\linewidth}{!}{
\begin{tabular}{l|c|c|c|c}\toprule
Model                                      &Type              &Param [B]               &Data size [M]         &ZS-FID $\downarrow$    \\ \midrule
DALL-E \cite{ramesh2021zero}               & AR               & 12                     & 250                  & 27.5                \\
Cogview  \cite{ding2021cogview}            & AR               & 4                      & 30                   & 27.1                 \\
Cogview2  \cite{ding2022cogview2}          & AR               & 6                      & 30                   & 24.0                 \\ 
Parti-350M   \cite{yu2022scaling}          & AR               & 0.35                   & \textgreater800      & 14.10                 \\ 
Parti-20B   \cite{yu2022scaling}           & AR               & 20                     & \textgreater800      & 7.23                 \\ \bottomrule
GLIDE  \cite{nichol2021glide}              & DF               & 5                      & 250                  & 12.24                 \\
\rowcolor{mygray}
LDM      \cite{rombach2022high}            & DF               & 1.45                   & 400                  & 12.63                 \\
DALL·E 2  \cite{ramesh2022hierarchical}    & DF               & 6.5                    & 250                  & 10.39                 \\
Imagen  \cite{saharia2022photorealistic}   & DF               & 7.9                    & 860                  & 7.27                 \\ 
eDiff-I  \cite{balaji2022ediffi}           & DF               & 9.1                    & 1000                 & 6.95                 \\ \bottomrule
LAFITE   \cite{zhou2022towards}            & GAN              & 0.15+0.08              & 3                    & 26.94                 \\
GALIP (CC3M)                               & GAN              & 0.24+0.08              & 3                    & 16.12                 \\
\rowcolor{mygray}
GALIP (CC12M)                              & GAN              & 0.24+0.08              & 12                   & 12.54        \\ \bottomrule
\end{tabular}}
\label{table2}
\vspace{-0.4cm}
\end{table}

Moreover, we evaluate the zero-shot text-to-image synthesis ability of our GALIP. The results are shown in Table \ref{table2}.
Compared with LAFITE \cite{zhou2022towards} trained on CC3M, our GALIP (CC3M) decreases FID from 26.94 to 16.12 significantly.
It demonstrates that integrating the CLIP model in the generator and discriminator is more effective than only introducing the CLIP loss for the GAN model.
Compared with autoregressive models (AR) and diffusion models (DF) which are pretrained with much larger model sizes and datasets, our GALIP also achieves competitive performance.
Especially, compared with LDM \cite{rombach2022high} which is one of the most important open-source large pretrained models, our GALIP achieves better performance even with much smaller model parameters and data size.
Furthermore, as shown in Figure~\ref{fig1}, our GALIP only requires 0.04s to generate one image which is $\sim$120$\times$faster than LDM \cite{rombach2022high}.
Besides, our GALIP can be inference on the CPU fastly without other acceleration settings.
This significantly reduces the hardware requirements of users.
In addition, the computational cost to pretrain our GALIP is quite less than these large pretrained autoregressive and diffusion models.
The GALIP of CC12M is only pretrained on 8$\times$3090 GPUs for 3 days. 
But these models require hundreds of GPUs and many weeks to pre-train.

\subsection{Qualitative Evaluation}
To evaluate the visual quality of synthesized images, we first compare the images synthesized by LAFITE \cite{zhou2022towards}, VQ-Diffusion \cite{gu2022vector}, and our GALIP which are trained on COCO in Figure~\ref{fig5}.
Then, we compare our GALIP (CC12M) with LDM (LAION-400M) \cite{rombach2022high,stablediff} in Figure~\ref{fig6}.

As shown in the 1$^{st}$, 2$^{nd}$, 4$^{th}$ and 5$^{th}$ columns of Figure~\ref{fig5}, the birds synthesized by LAFITE \cite{zhou2022towards} and VQ-Diffusion \cite{gu2022vector} contain break or wrong shapes. 
Moreover, both LAFITE \cite{zhou2022towards} and VQ-Diffusion \cite{gu2022vector} lose some fine-grained visual features (e.g., 1$^{st}$, 2$^{nd}$, 5$^{th}$ and 6$^{th}$ columns), which makes the synthesized images lack details and look unreal.
However, the images synthesized by our GALIP have correct object shapes and clear fine-grained contents.

The superiority is more obvious in complex COCO images, which contain various shapes and multiple objects.
As the results are shown in the 7$^{th}$, 8$^{th}$, 9$^{th}$, 10$^{th}$ columns of Figure~\ref{fig6}, the LAFITE \cite{zhou2022towards} and VQ-Diffusion \cite{gu2022vector} models cannot synthesize the right shape of ``train'', ``children'', ``woman'', and ``stuffed bear''.
Furthermore, they also cannot synthesize the right visual concept of ``showing off toy cell phone'' and ``sitting on a book shelf''.
However, armed with the proposed CLIP-based D and CLIP-empowered G, our GALIP can cope with more strict visual requirements and synthesize various shapes of different objects (see 8$^{th}$, 9$^{th}$, 10$^{th}$ and 12$^{th}$ columns) and present the right visual concepts in synthesized images.
We also observe that LAFITE \cite{zhou2022towards} and VQ-Diffusion \cite{gu2022vector} also can not synthesize correct human facial features.
For example, as shown in the 8$^{th}$, 9$^{th}$, 12$^{th}$, they can not synthesize realistic human faces.
But our GALIP can synthesize these features correctly.

Moreover, we compare the images synthesized by the LDM (LAION-400M) \cite{rombach2022high,stablediff} and our GALIP (CC12M) in Figure~\ref{fig6}.
As the results are shown in the 1$^{st}$, 4$^{th}$, 5$^{th}$, 8$^{th}$, 11$^{th}$ columns of Figure~\ref{fig6},
the LDM does not generate the objects (``ghost'', ``teddy bear'', ``modem'', ``person'', ``model'') described in the texts,
but our GALIP can synthesize these objects correctly.
Also, our model can generate correct visual features such as ``shining eyes'', ``Blue Lighthouse'', ``smiling statue'', and ``surprised girl'' in the 3$^{rd}$, 6$^{th}$, 7$^{th}$, 10$^{th}$ columns.
Furthermore, as shown in the 9$^{th}$, 10$^{th}$, and 12$^{th}$ columns of Figure~\ref{fig6}, our GALIP keeps the superior performance of human face synthesis.
The extensive quantitative evaluation results demonstrate the superiority and effectiveness of our proposed GALIP, which is able to generate high-fidelity, creative and complex images with various shapes and multiple objects.

\begin{figure}[t] \small
  \centering
  \includegraphics[width=\linewidth]{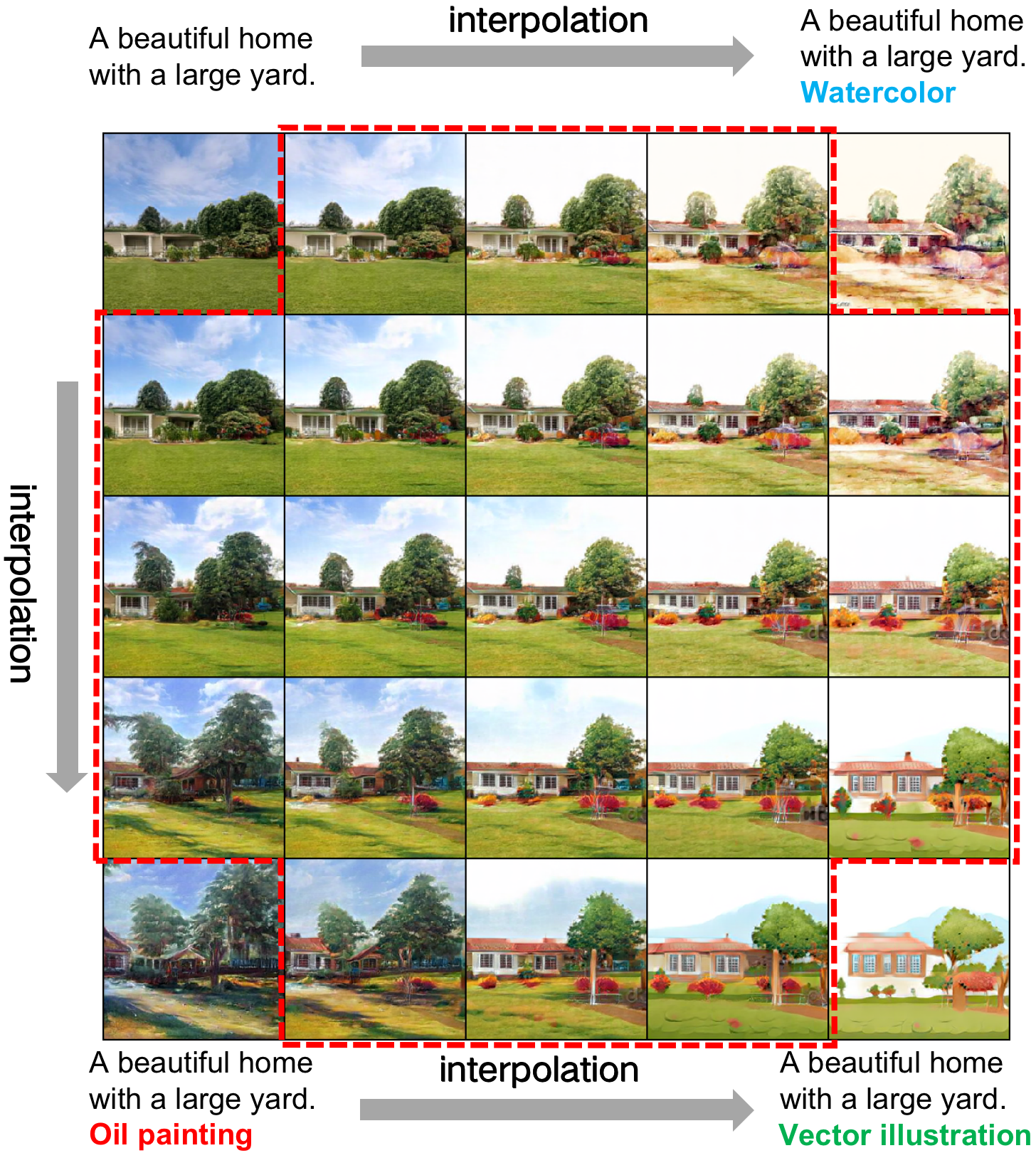}
  \caption{Images synthesized by interpolating four-sentence embeddings. Our GALIP supports gradual changes when interpolating sentence embeddings describing different image styles. It makes the degree of stylization of the image controllable and creates new styles by blending different styles.}
  \label{fig7}
  \vspace{-0.4cm}
\end{figure}

Additionally, we conduct some experiments to show the smooth latent space of our GALIP.
Current autoregressive and diffusion models are sensitive to input sentences.
This instability makes users need to try a lot of prompts to get satisfied images.
Differently, our GALIP inherits the smooth latent space from GAN, it enables gradual and smooth changes along with text changes.
As shown in Figure~\ref{fig7}, there is a smooth transition of synthesized images from top to bottom, left to right. 
The smooth latent space makes the degree of stylization of the image controllable.
The users can fine-tune synthesized image styles like a style knob, and it also enables the users to create new styles by blending different image styles, as highlighted by the red dashed lines.

\subsection{Ablation Study}

To verify the effectiveness of each component in the proposed GALIP, we conduct ablation studies on the test set of the COCO dataset.
The components being evaluated in this subsection include CLIP-based D (CD) and CLIP-empowered G (CG). 
We also further conduct ablation studies on Bridge-FP (BFP) and Prompt Predictor (PP) in CLIP-empowered G, and CLIP-FE (CFE) in CLIP-based D.
Furthermore, we compare our CLIP-FE with CCM\&CSM of Projected GAN \cite{sauer2021projected}, which yields a U-Net architecture to enable multi-scale feedback. 
In addition, we investigate the layer choice strategy of CLIP-FE and Prompt Predictor. 
The results on the COCO dataset are shown in Table~\ref{table3}.

\begin{table}[t] \small
\centering
\caption{The performance of different components of our model on the test set of COCO.}
\begin{tabular}{l|c|c}\toprule
Architecture                                              &FID $\downarrow$             &CS $\uparrow$ \\ \midrule
Baseline                                                  & 17.31                       & 0.2996         \\
Baseline w/ CD w/ CFE                                     & 7.92                        & 0.3221          \\
Baseline w/ CD w/ CCM\&CSM                                & 10.77                       & 0.3123           \\
Baseline w/ CD w/ BFP                                     & 6.52                        & 0.3301           \\
Baseline w/ CD w/ BFP w/ PP (GALIP)                       & \textbf{5.85}               & \textbf{0.3338}\\ \bottomrule
GALIP w/ CFE (2$^{nd}$)                                   & 13.41                 & 0.3015          \\
GALIP w/ CFE (5$^{th}$)                                  & 8.60                   & 0.3145          \\
GALIP w/ CFE (12$^{th}$)                                 & 10.72                  & 0.3104          \\
GALIP w/ CFE (2$^{nd}$,5$^{th}$)                         & 6.70                   & 0.3301          \\
GALIP w/ CFE (2$^{nd}$,5$^{th}$,12$^{th}$)               & 6.61                   & 0.3305          \\
GALIP w/ CFE (2$^{nd}$,5$^{th}$,9$^{th}$)                & \textbf{5.85}          & \textbf{0.3338}          \\
GALIP w/ CFE (2$^{nd}$,5$^{th}$,8$^{th}$,9$^{th}$)       & 6.01                   & 0.3305          \\\bottomrule

GALIP w/ PP (1$^{st}$-12$^{th}$)                          & 6.24                  & 0.3320           \\
GALIP w/ PP (1$^{st}$-9$^{th}$)                           & \textbf{5.85}         & \textbf{0.3338}          \\
GALIP w/ PP (1$^{st}$-6$^{th}$)                           & 6.40                  & 0.3310          \\
GALIP w/ PP (1$^{st}$-3$^{th}$)                           & 6.52                  & 0.3305          \\\bottomrule
\end{tabular}
\label{table3}
\vspace{-0.4cm}
\end{table}

\noindent{\bf Baseline.} Our baseline is a one-stage text-to-image GAN \cite{tao2020df}.
It is composed of a CLIP text encoder and CNN-based generator and discriminator.
And it generates complex images from sentence vectors directly.

\noindent{\bf Effect of CLIP-based D and CLIP-FE.} The CLIP-based D decreases FID from 17.31 to 7.92 and improves CLIMSIM (CS) from 0.2996 to 0.3221. 
The results demonstrate that the complex scene understanding ability of CLIP-ViT promotes the complex image synthesis ability significantly.
Furthermore, we compared our CLIP-FE (CFE) with CCM\&CSM \cite{sauer2021projected}. 
Our CLIP-FE achieves better FID and CLIPSIM.
It shows that our CLIP-FE is more effective in extracting informative visual features from CLIP-ViT.

\noindent{\bf Effect of CLIP-empowered G and Bridge-FP.} The CLIP-empowered G with Bridge-FP further decreases FID from 7.92 to 6.52 and improves CLIPSIM from 0.3221 to 0.3301. 
It demonstrates that predicted bridge features and CLIP-ViT can enhance the complex image synthesis ability effectively.

\noindent{\bf Effect of Prompt Predictor.} The proposed Prompt Predictor (PP) also decreases FID from 6.52 to 5.85 and improves CLIPSIM from 0.3301 to 0.3338. 
The result demonstrates that the Prompt Predictor makes the CLIP-ViT more suitable for generation tasks and induces more meaningful features from CLIP-ViT to improve the generative ability.

\noindent{\bf CLIP Layer Selection.} We find that the last few layers of CLIP-ViT defect the performance of CLIP-based D.
The reason may be that the first layers of CLIP-ViT extract useful visual features and understand complex images, 
and the last layers focus on generalization ability to align with high-level concepts in text features.
The generalization ability may defect the performance of CLIP-based D because it reduces the differences between synthetic and real images and weakens the discriminator.
Conversely, since CLIP-empowered G requires the generalization ability to map the bridge feature to meaningful visual features, adding prompts in the last few layers may defect the generalization ability.
So we extract the CLIP features from 2$^{nd}$,5$^{th}$,and 9$^{th}$ layers in CLIP-based D, and add prompts to 1$^{st}$-9$^{th}$ layers.
And we find that extracting more CLIP features does not lead to better performance.

\begin{figure}[t] \small
  \centering
  \includegraphics[width=\linewidth]{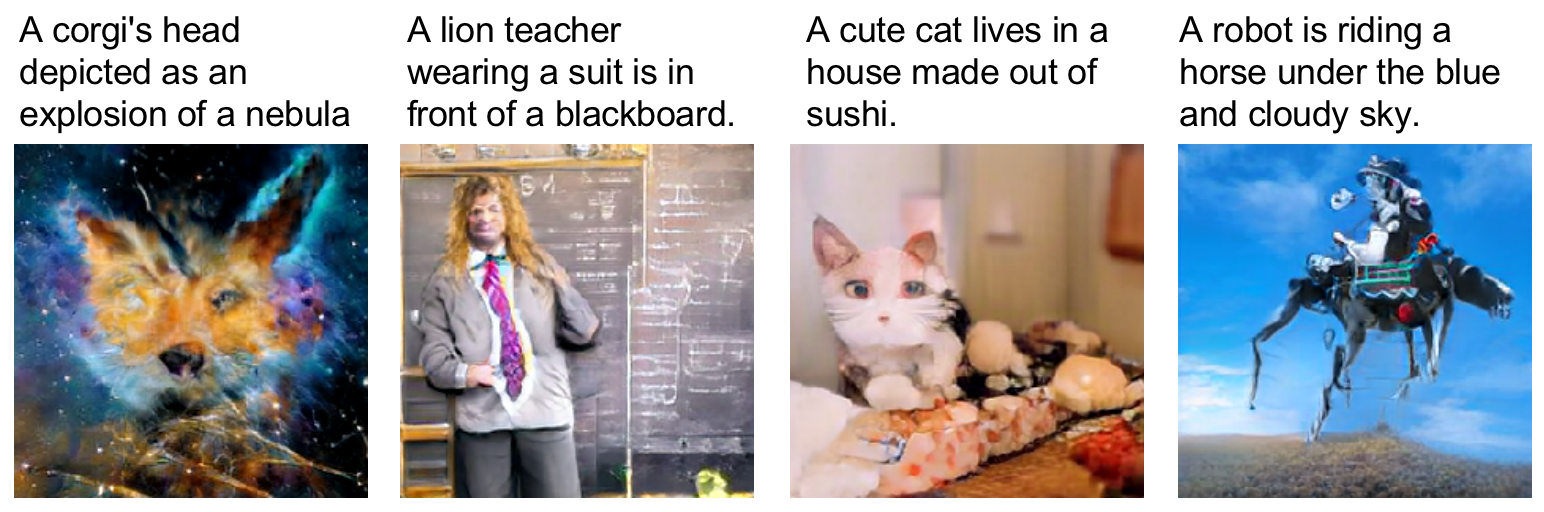}
  \caption{Illustration of failure cases. It is still hard for current GALIP to synthesize some imaginary images. Enlarging the model size and training data may improve image quality.}
  \label{fig8}
  \vspace{-0.4cm}
\end{figure}

\subsection{Limitations}
Our GALIP shows superiority in text-to-image synthesis, but some limitations should be considered in future studies. 
First, our model employs the CLIP model to provide text features for the generator and discriminator.
However, current models \cite{saharia2022photorealistic} show that the generic large language models \cite{raffel2020exploring} (e.g., T5) improve the performance of text-to-image synthesis effectively.
Replacing the CLIP text encoder with T5 may further improve the performance.
Second, the model size and pretraining dataset are much smaller than other large pretrained models \cite{rombach2022high,yu2022scaling,ramesh2022hierarchical,saharia2022photorealistic,balaji2022ediffi}, 
it limits the synthesis ability of imaginary images (see Figure~\ref{fig8}). 
Pretraining on a larger dataset  with a larger model size may benefit the performance.
We will try to address these limitations in our future work.

\section{Conclusion}
In this paper, we propose a novel Generative Adversarial CLIPs (GALIP) for text-to-image synthesis. 
Compared with previous models, our GALIP can synthesize higher-quality complex images.
Moreover, we propose a CLIP-based discriminator and CLIP-empowered generator, which exerts the complex scene understanding and domain generalization ability of CLIP.
Our GALIP achieves significant improvements on challenging datasets.
Furthermore, current large models are pretrained for generative or understanding tasks.
In this work, we integrate the understanding model (CLIP-ViT) into a generative model and achieve impressive results.
It shows that there are some commonalities between understanding and generative models.
This may be enlightening for building a general large model.

{\small
\bibliographystyle{ieee_fullname}
\bibliography{egbib}
}

\end{document}